%% file: Template.tex
\documentclass{article}
\usepackage{spconf,amsmath,graphicx,hyperref}
\usepackage{booktabs}   
\usepackage{colortbl}   
\arrayrulecolor{black}  
\usepackage{cite}
\usepackage{booktabs,pifont}
\usepackage{url}
\usepackage{hyperref}
\usepackage{amssymb}  

\newcommand{\cmark}{\ding{51}} 

\newcommand{\method}{\textsc{Mem4Teeth}} 
\graphicspath{{figs/}{figures/}}

\title{\method: Memory-Guided Point Cloud Completion for Dental Reconstruction}

\name{Jianan Sun$^{\star}$, Yukang Huang$^{\star}$, Dongzhihan Wang$\dagger$, Mingyu Fan$^{\star}$} 
\address{$^{\star}$ College of Information and Intelligent Science, Donghua University, Shanghai, China\\
$^{\dagger}$ Institute of Artificial Intelligence, Shanghai University, Shanghai, China \\ 
fanmingyu@dhu.edu.cn}

\begin{document}
\maketitle

\begin{abstract}
Partial dental point clouds often suffer from large missing regions caused by occlusion and limited scanning views, which bias encoder-only global features and force decoders to hallucinate structures. We propose a retrieval-augmented framework for tooth completion that integrates a prototype memory into standard encoder--decoder pipelines. After encoding a partial input into a global descriptor, the model retrieves the nearest manifold prototype from a learnable memory and fuses it with the query feature through confidence-gated weighting before decoding. The memory is optimized end-to-end and self-organizes into reusable tooth-shape prototypes without requiring tooth-position labels, thereby providing structural priors that stabilize missing-region inference and free decoder capacity for detail recovery. The module is plug-and-play and compatible with common completion backbones, while keeping the same training losses. Experiments on a self-processed Teeth3DS benchmark demonstrate consistent improvements in Chamfer Distance, with visualizations showing sharper cusps, ridges, and interproximal transitions. Our approach provides a simple yet effective way to exploit cross-sample regularities for more accurate and faithful dental point-cloud completion.
\end{abstract}

\textbf{Code:} \url{https://github.com/flyingfish118/Memory-Guided-Point-Cloud-Completion-for-Dental-Reconstruction}

\begin{keywords}
Point cloud completion, dental reconstruction, memory bank, retrieval and fusion
\end{keywords}

\section{Introduction}
\label{sec:intro}
\input{sections/01_introduction}

\section{Method}
\label{sec:method}
\input{sections/02_method}

\section{Experiments}
\label{sec:experiments}
\input{sections/03_experiments}

\section{Conclusion}
\label{sec:conclusion}
\input{sections/04_conclusion}

\bibliographystyle{IEEEbib}
\bibliography{refs/refs}

\end{document}

%% file: sections/01_introduction.tex
\begin{figure*}[t]
  \centering
  \includegraphics[width=0.85\textwidth]{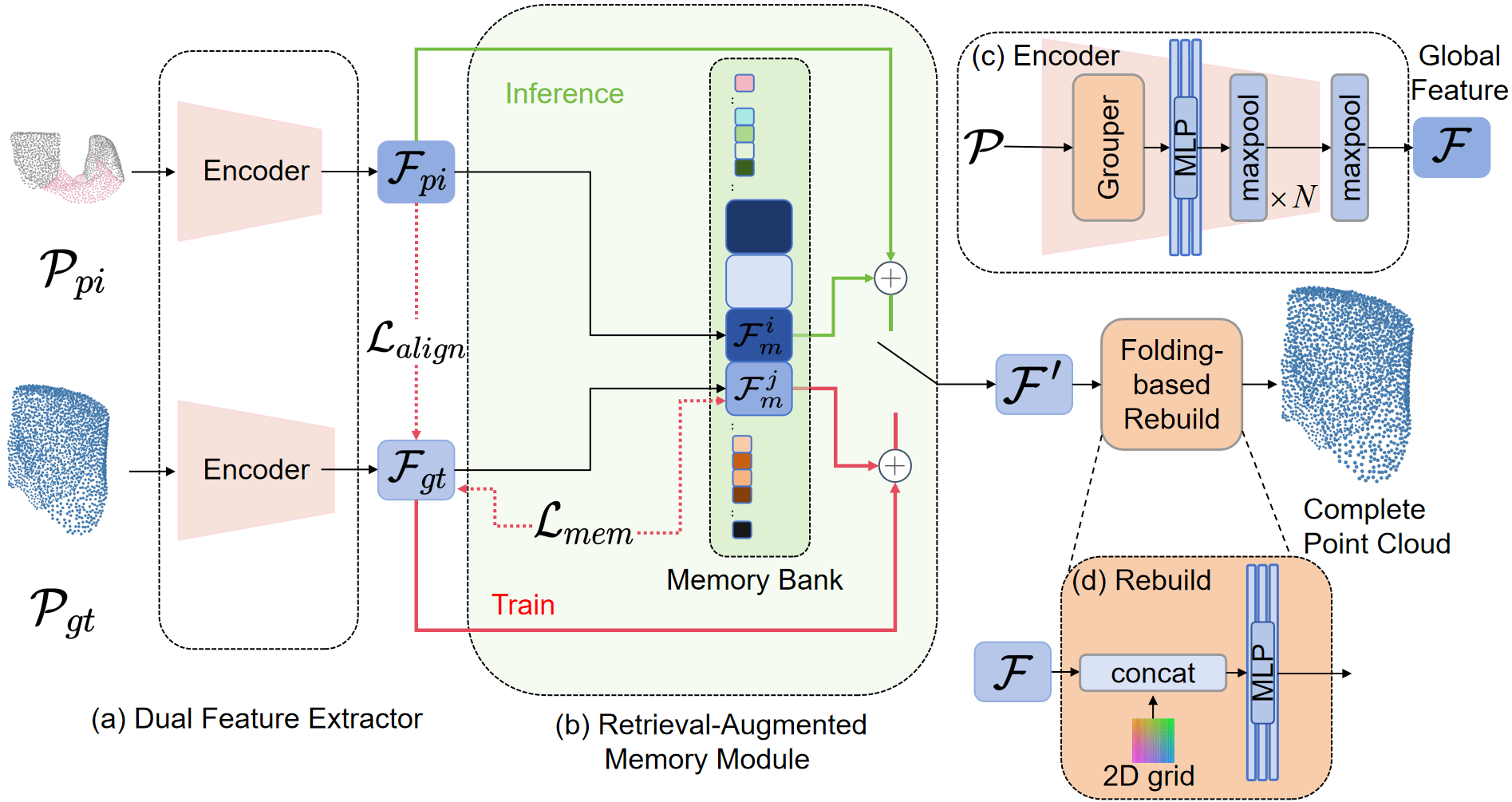}
\caption{
Overview of our retrieval-augmented completion framework.
\textbf{(a) Dual Feature Extractor:} Two encoders with identical architecture but no weight sharing produce global descriptors for the partial input ($\mathcal{F}_{pi}$) and complete ground truth ($\mathcal{F}_{gt}$).
\textbf{(b) Retrieval-Augmented Memory Module:} A learnable prototype bank $\{F_m^k\}_{k=1}^K$ is queried by features. During \emph{training}, $\mathcal{F}_{gt}$ selects its nearest prototype $F_m^{j}$ to update the prototype memory ($\mathcal{L}_{\text{code}}$) and align distributions with the partial branch ($\mathcal{L}_{\text{align}}$). During \emph{inference}, $\mathcal{F}_{pi}$ retrieves $F_m^{i}$; the prototype is confidence-gated and fused to yield the debiased global $\mathcal{F}'$.
\textbf{(c--d)} We use a standard encoder backbone and a folding-based decoder to generate the completed point cloud from $\mathcal{F}'$.
}
\label{fig:arch}
\end{figure*}

Dental 3D scans frequently yield partial tooth point clouds due to occlusion, limited viewpoints, and reflectance artifacts\cite{yuan2018pcn}\cite{wei2020tanet}. Under such incompleteness, encoder–decoder completion networks must infer large missing regions from a biased global representation distilled from the visible subset\cite{zhu2025pointsea}. This bias propagates to the decoder, which is then forced to “hallucinate” structure with weak guidance, often leading to mode averaging, over-smoothed crowns, and unstable reconstruction of clinically salient details such as cusps, ridges, and interproximal surfaces\cite{hosseinimanesh2023mesh}\cite{ziruo2023variational}.

Existing approaches mitigate the ill-posedness mainly by strengthening the backbone (e.g., deeper transformers)\cite{vaswani2017attention}\cite{ziruo2023variational}\cite{broll2024generative}\cite{fan2024collaborative}, adding local geometric constraints, or injecting generic priors through losses or diffusion-style decoders\cite{hosseinimanesh2023mesh}\cite{yang2024dcrownformer}\cite{ding2023morphology}\cite{tian2022efficient}. However, they largely ignore cross-sample shape regularities that are pronounced in dentistry: many teeth share homologous meso-scale geometry that recurs across subjects. Methods that do use priors typically require explicit labels (e.g., tooth position) or handcrafted templates, which are costly, brittle across datasets, and risk leaking category bias. Moreover, stronger generative priors increase compute and training complexity without guaranteeing that the inferred global feature is corrected toward a plausible manifold point for the specific partial input\cite{yu2021pointr}\cite{li2023proxyformer}.

What is missing is a mechanism that actively consults reusable shape knowledge at the moment the biased global feature is formed\cite{gao2023retrieval}. Instead of relying solely on the current partial to define the completion space, the model should retrieve a compact, discriminative prototype from past shapes that lie on the same manifold neighborhood, then use it to de-bias the global representation before decoding. Such retrieval-augmented correction would (i) provide structural anchors for large missing regions, (ii) free decoder capacity for fine-detail synthesis.

We introduce a retrieval-augmented \emph{prototype memory} for dental point-cloud completion. The encoder’s global feature queries a compact bank of learnable manifold prototypes; the retrieved prototype then \emph{debiases} the global representation via confidence-gated fusion before decoding. Without tooth-position labels, the memory self-organizes to capture strong cross-sample dental regularities, providing stable structural priors and enabling sharper detail recovery.

\noindent\textbf{In summary, our contributions are threefold:}
\begin{itemize}
\item We propose a plug-and-play prototype memory that debiases encoder global features through similarity-based retrieval and confidence-gated fusion.
\item We develop a geometric constraints-free prototype learning mechanism in which manifold prototypes self-organize to reflect dental shape regularities and supply structural anchors for large missing regions.
\item We demonstrate consistent gains in completeness and fine geometry across backbones with minimal overhead.
\end{itemize}

%% file: sections/02_method.tex
\subsection{Overview}
Given a partial tooth point cloud $\mathcal{P}_{pi}\!\in\!\mathbb{R}^{N\times 3}$ and its complete ground truth $\mathcal{P}_{gt}\!\in\!\mathbb{R}^{N\times 3}$, our model aims to predict a completed point cloud $\widehat{\mathcal{P}}$ that preserves global structure while restoring dental fine details. 
Figure~\ref{fig:arch} illustrates our pipeline: dual encoders extract global descriptors, a retrieval-augmented prototype memory provides structural anchors, and a decoder reconstructs the final output.

\subsection{Dual Feature Extractor}
We adopt two encoders of identical architecture but without weight sharing.  
The partial encoder $f_{\theta}$ maps $\mathcal{P}_{pi}$ to a global descriptor $\mathcal{F}_{pi}=f_{\theta}(\mathcal{P}_{pi})\in\mathbb{R}^{d}$, while the ground-truth encoder $f_{\phi}$ maps $\mathcal{P}_{gt}$ to $\mathcal{F}_{gt}=f_{\phi}(\mathcal{P}_{gt})$.  
During training, $\mathcal{F}_{gt}$ serves as a reliable reference less biased by missing regions. At inference, only $\mathcal{F}_{pi}$ is available.

\subsection{Prototype Memory Bank}
We maintain a prototype memory $\mathcal{M}=\{F_m^1,F_m^2,\dots,F_m^K\}$, where each prototype $F_m^k \in \mathbb{R}^d$ represents a manifold mode learned across training samples.  
Instead of projecting features into an auxiliary space, we directly retrieve the \emph{nearest} prototype to a given feature vector:
\begin{align}
F_m^j &= \arg\min_{F_m^k \in \mathcal{M}} \| \mathcal{F}_{gt} - F_m^k \|_2^2, \\
F_m^i &= \arg\min_{F_m^k \in \mathcal{M}} \| \mathcal{F}_{pi} - F_m^k \|_2^2.
\end{align}
Here $F_m^j$ is the closest prototype to the GT descriptor, and $F_m^i$ is the closest prototype to the partial descriptor.

During training, we fuse the GT feature $\mathcal{F}_{gt}$ with its retrieved prototype $F_m^j$ to form a debiased descriptor:
\begin{equation}
\mathcal{F}' = (1-\alpha)\,\mathcal{F}_{gt} + \alpha\,F_m^j, \qquad \alpha \in [0,1].
\end{equation}
The fusion weight $\alpha$ is controlled by a confidence estimator (e.g., similarity score or entropy). This design drives the prototype memory to align with complete-shape features and provides stable structural anchors.

At inference, when $\mathcal{F}_{gt}$ is unavailable, we use the partial descriptor and its nearest prototype:
\begin{equation}
\mathcal{F}' = (1-\alpha)\,\mathcal{F}_{pi} + \alpha\,F_m^i.
\end{equation}
This debiases the partial feature towards a plausible manifold neighborhood while retaining instance-specific cues.

\subsection{Decoder}
For point cloud reconstruction, we adopt a folding-based decoder following FoldingNet\cite{yang2018foldingnet}. 
Starting from a fixed 2D grid, the decoder conditions each grid point on the fused global feature $\mathcal{F}'$ and progressively ``folds'' the grid into 3D space through lightweight MLPs. 
This design provides a simple and effective way to generate the completed point cloud $\widehat{\mathcal{P}}$.

\subsection{Loss Functions}
We retain two key objectives to train the prototype memory and align distributions:
To update prototypes with reliable GT descriptors, we adopt a VQ-style commitment loss:
\begin{equation}
\mathcal{L}_{\text{mem}} 
= \big\|\mathrm{sg}(\mathcal{F}_{gt}) - F_m^j\big\|_2^2
+ \big\|\mathcal{F}_{gt} - \mathrm{sg}(F_m^j)\big\|_2^2,
\end{equation}
where $\mathrm{sg}(\cdot)$ denotes the stop-gradient operator. This ensures prototypes remain faithful representatives of GT-distribution features.
We also encourage consistency between partial and GT encoders via:
\begin{equation}
\mathcal{L}_{\text{align}} = \|\mathcal{F}_{pi} - \mathcal{F}_{gt}\|_2^2,
\end{equation}
or an InfoNCE variant to enforce similarity of positive pairs and separation from negatives.
The decoder $D(\cdot)$ maps $\mathcal{F}'$ to $\widehat{\mathcal{P}}=D(\mathcal{F}')$. We adopt Chamfer distance\cite{fan2017point} and F-score as in prior work:
\begin{equation}
\mathcal{L}_{\text{cd}} = \mathrm{CD}(\widehat{\mathcal{P}},\mathcal{P}_{gt}).
\end{equation}
The full loss is a weighted combination:
\begin{equation}
\mathcal{L} = \mathcal{L}_{\text{cd}} + \lambda_f \mathcal{L}_{\text{f}}
+ \lambda_{\text{align}} \mathcal{L}_{\text{align}}
+ \lambda_{\text{mem}} \mathcal{L}_{\text{mem}}.
\end{equation}

\subsection{Inference and Complexity}
At test time, only the partial encoder, memory retrieval, and decoder are active. Retrieval is $O(Kd)$ per query with $K\!<\!128$, yielding negligible overhead. The design is plug-and-play and does not rely on tooth-position labels, letting prototypes self-organize into reusable manifold anchors.

%% file: sections/03_experiments.tex
\subsection{Datasets and Evaluation Metrics}
We evaluate on a self-processed dataset derived from \textit{Teeth3DS}, which provides tooth-position and gingiva segmentation but no completion pairs. For each case, we extract mesh vertices and construct a pair \((\mathcal{P}_{pi},\mathcal{P}_{gt})\): the target tooth vertices form the complete ground truth \(\mathcal{P}_{gt}\), while the partial \(\mathcal{P}_{pi}\) is obtained by removing this tooth and retaining its local context (adjacent teeth and the \(n\) nearest gingiva points). After zero-mean/unit-scale normalization, the partial is resampled to 2048 points using farthest-point sampling or duplication. During training, \((\mathcal{P}_{pi},\mathcal{P}_{gt})\) are processed by dual encoders to yield \((\mathcal{F}_{pi},\mathcal{F}_{gt})\); the prototype memory is updated with \(\mathcal{F}_{gt}\), and the fused feature \(\mathcal{F}'\) is decoded. Supervision employs the symmetric Chamfer distance (CD-L2$\downarrow$)~\cite{fan2017point} between prediction \(\widehat{\mathcal{P}}\) and ground truth \(\mathcal{P}_{gt}\). At inference, \(\widehat{\mathcal{P}}\) is generated from \(\mathcal{P}_{pi}\), and for visualization we concatenate it with the retained context. Prototypes self-organize without tooth-position labels, and all methods share identical preprocessing, normalization, and train/val/test splits for fair comparison.

\noindent\textbf{Evaluation Metrics.} We use a standard metrics to assess completion quality:
\begin{itemize}
\item \textbf{Chamfer Distance (CD):} Average closest-point distance between predicted and ground-truth point sets.
\end{itemize}

\subsection{Comparison Results}
We compare our method against four representative completion baselines—\textit{FoldingNet}\cite{yang2018foldingnet}, \textit{PCN}\cite{yuan2018pcn}, \textit{PoinTr}\cite{yu2021pointr}, and \textit{SVDFormer}\cite{zhu2023svdformer}. 
All models are trained and evaluated on the same splits of our Teeth3DS-derived dataset with identical preprocessing, point budgets, and metrics. 
We report symmetric Chamfer Distance (CD-L2$\downarrow$) averaged over the test set, with implementations following the official code of each baseline. 
As summarized in Table~\ref{tab:cmp_main}, our retrieval-augmented prototype memory achieves the best overall accuracy (numbers in \textbf{bold}), reflecting stronger global consistency and finer local geometry. 
Quantitatively, our method reduces CD-L2 by \textbf{15.7\%} compared with the current state-of-the-art SVDFormer, and by \textbf{58.5\%} compared with the classical PCN baseline.

\input{tables/comparison.tex}

\subsection{Qualitative Visualization}
\begin{figure}[t]
\centering
\includegraphics[width=\linewidth]{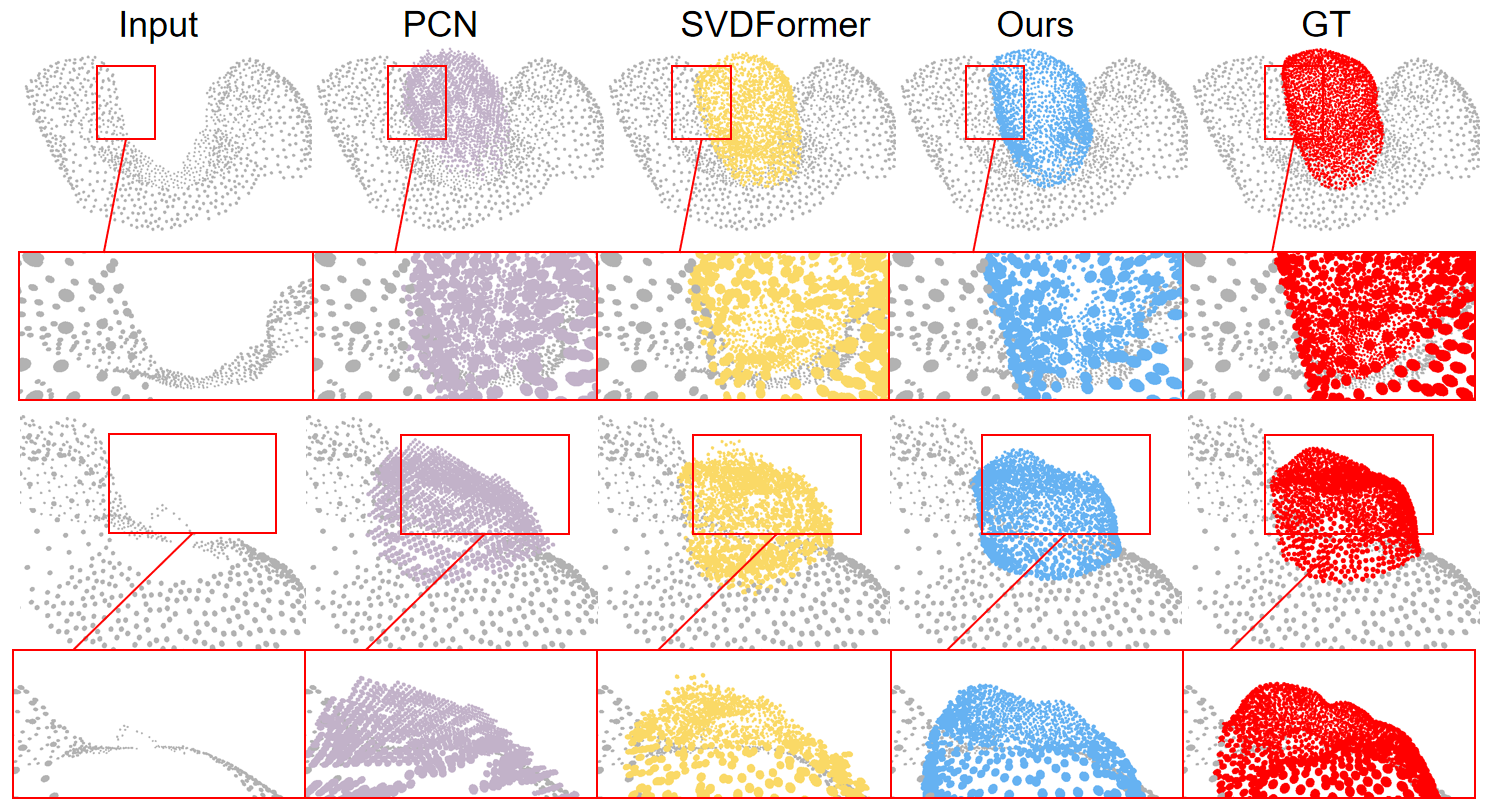}
\caption{Qualitative dental completion results: $\mathcal{P}_{pi}$ (left), $\widehat{\mathcal{P}}$ (middle), $\mathcal{P}_{gt}$ (right).}
\label{fig:qual}
\end{figure}

Figure~\ref{fig:qual} compares completions on incisors (top row) and molars (bottom row). 
For incisors, baselines frequently overfill the interproximal region and penetrate neighboring teeth (see red boxes), producing visible tooth–tooth collisions; our method preserves the interdental space and yields a clean crown boundary, avoiding overlap with the adjacent tooth. 
For molars, baselines struggle to reconstruct the occlusal surface, leading to flattened or noisy cusps and missing pits/fissures, whereas our result recovers a coherent occlusal pattern with sharper cusp tips and continuous ridges. 
These improvements stem from retrieving a prototype from the memory as a structural prior and fusing it with the partial feature, which stabilizes large missing regions while retaining instance-specific details.

\subsection{Prototype Evidence via UMAP}
\begin{figure}[t]
\centering
\includegraphics[width=\linewidth]{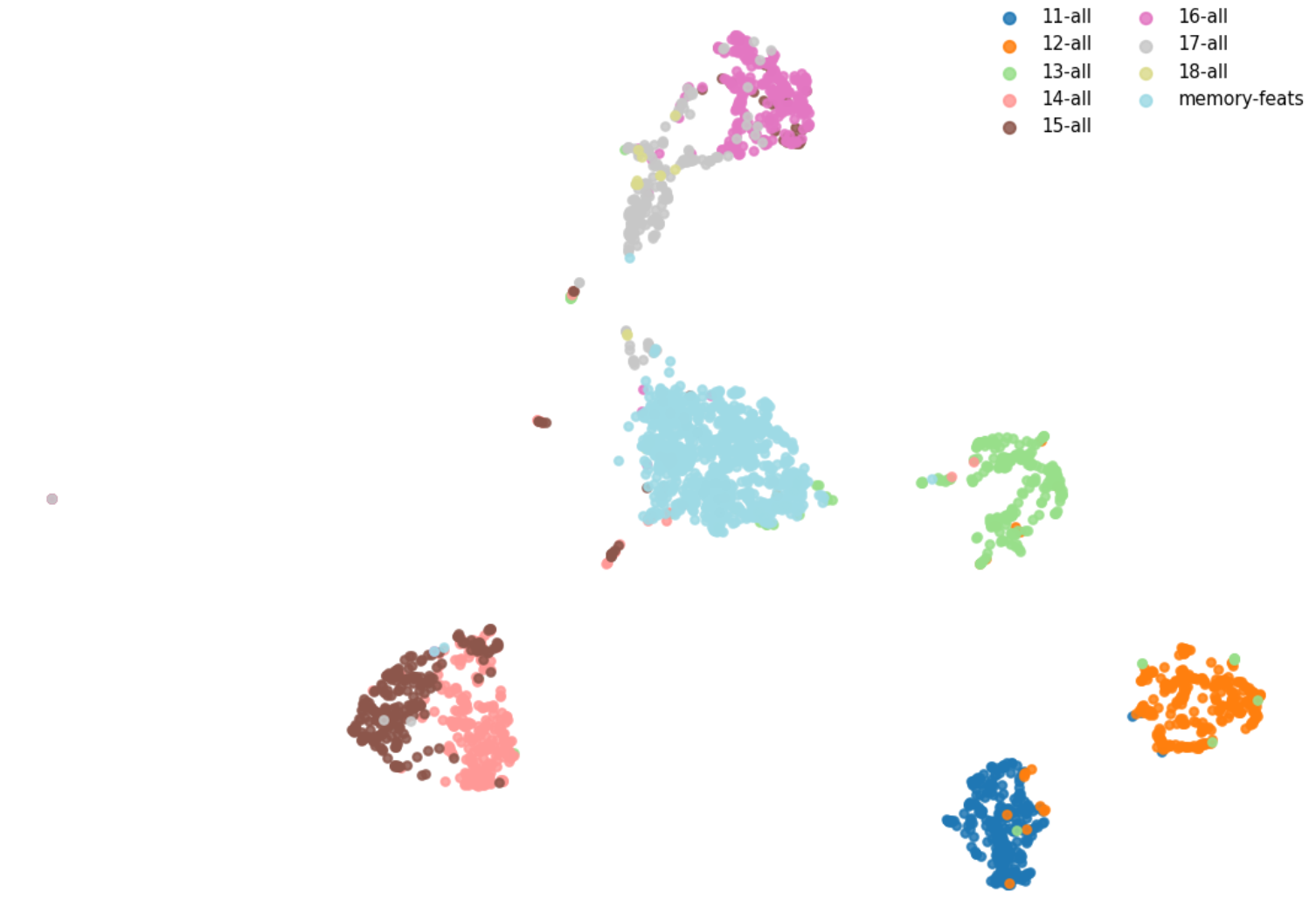}
\caption{UMAP of encoder features and prototypes. }
\label{fig:umap}
\end{figure}

The UMAP in Fig.~\ref{fig:umap} provides evidence that the proposed memory acts as a canonical prior rather than a copy of the partial-feature distribution. The blue points (memory features) form a compact, high-density cluster, indicating that the learned prototypes converge to a low-variance, denoised region of the manifold. In contrast, the colored points (partial encoder features) split into position-dependent groups that reflect anatomical variability and visibility patterns in partial scans. Importantly, these groups are not random: central vs. lateral incisors (e.g., 11/12) are separable yet lie in close proximity, consistent with their morphological continuity; the canine (13) forms a distinct cluster, which accords with its singular crown geometry; the first and second premolars (14/15) partially mix while remaining sub-separable, and the molars (16/17/18) coalesce into a broader cluster due to shared multi-cusp topology. This structured arrangement supports our hypothesis that partial features are biased by tooth position and occlusion, whereas the memory summarizes cross-sample regularities into a shared anchor space. Retrieval from this compact memory, followed by confidence-gated fusion, therefore pulls biased partial descriptors toward a canonical region while preserving instance-specific cues—precisely the behavior required for stabilizing large missing-region inference. Note that UMAP preserves local neighborhoods but not global distances; the compact memory cluster alongside anatomically meaningful partial clusters is thus a desirable signature of a prior–observation decomposition, and empirically aligns with the observed gains in CD-L2.

\subsection{Ablation Studies}
We ablate two components: Prototype Memory (PM) and Dual Encoders (DE; no weight sharing). 
Table~\ref{tab:ablate_pm_de} toggles these modules. Removing PM loses structural anchors for large missing regions; removing DE (i.e., sharing encoder weights) weakens distribution alignment between partial and GT branches. The full model (PM+DE) achieves the lowest CD-L2.
\input{tables/tab_ablation}

%% file: tables/comparison.tex
\begin{table}[t]
\centering
\caption{Comparison on our Teeth3DS-derived test set. Lower is better for CD-L2 and higher is better for F-score@1\%.}
\label{tab:cmp_main}
\setlength{\tabcolsep}{10pt}
\begin{tabular}{lcc}
\toprule
\textbf{Methods} & \textbf{CD-L2} ($\times 10^{-4}$) $\downarrow$\\
\midrule
FoldingNet & 5.61 \\
PCN & 3.76  \\
PoinTr & 2.43  \\
SVDFormer & 1.85  \\
\midrule
\textbf{Ours} & \textbf{1.56} \\
\bottomrule
\end{tabular}
\end{table}

%% file: tables/tab_ablation.tex
\begin{table}[t]
\centering
\caption{Ablation on Prototype Memory (PM) and Dual Encoders (DE) 
on our Teeth3DS-derived test set (Chamfer Distance, CD-L2 $\downarrow$).}
\label{tab:ablate_pm_de}
\setlength{\tabcolsep}{10pt}
\renewcommand{\arraystretch}{1.05}
\begin{tabular}{ccc}
\toprule
\textbf{PM} & \textbf{DE} & \textbf{CD-L2} ($\times 10^{-4}$) $\downarrow$ \\
\midrule
--      & --      & 2.43 \\
\cmark  & --      & 1.92 \\
\addlinespace[2pt]
\textbf{\cmark} & \textbf{\cmark} & \textbf{1.56} \\
\bottomrule
\end{tabular}
\end{table}

%% file: sections/04_conclusion.tex

We presented \method, a memory-guided framework for dental point cloud completion that retrieves anatomically similar prototypes and fuses them with latent features to enhance occlusal details and crown consistency. On Teeth3DS, \method achieves consistent CD-$\ell_2$ gains with clear improvements on clinically relevant structures. However, performance relies on memory coverage, and rare cases or distribution shifts may lower retrieval quality, while the memory index adds modest storage and lookup cost. In future work, we plan to integrate lightweight diffusion refinement, adopt uncertainty-driven retrieval to down-weight unreliable partials, and extend the memory to cross-modality priors for stronger anatomy constraints.